\UseRawInputEncoding
\documentclass[10pt,twocolumn,letterpaper]{article} 

\usepackage{avss}
\usepackage{times}
\usepackage{epsfig}
\usepackage{graphicx}
\usepackage{amsmath}
\usepackage{amssymb}
\usepackage{tabularx}
\usepackage{xcolor}
\usepackage{bbm}
\newif\ifcomments
\commentstrue
\commentsfalse

\ifcomments
    \usepackage{ulem}
    \def\mc#1{{\color{blue} [\textbf{MC:} #1]}}
    
    \def\jlk#1{{\color{red} [\textbf{JLK:} #1]}}
    
    \def\picomment#1{{$\!$\color{magenta} [PI: #1]}}

\else 
    \def\mc#1{}
    
    \def\jlk#1{}
    
    \def\picomment#1{}
    
    \def\jbedit#1{}
    
\fi

\usepackage[pagebackref=true,breaklinks=true,letterpaper=true,colorlinks,bookmarks=false]{hyperref}

\avssfinalcopy % *** Uncomment this line for the final submission

\def\avssPaperID{35} % *** Enter the AVSS Paper ID here

\ifavssfinal\fi
\begin{document}

%%%%%%%%% TITLE
\title{FRIDA: Fisheye Re-Identification Dataset with Annotations}

\author{Mertcan Cokbas, John Bolognino, Janusz Konrad, Prakash Ishwar\thanks{This work was supported by ARPA-E (agreement DE-AR0000944) and by Boston University Undergraduate Research Opportunities Program.
% 978-1-6654-6382-9/22/\$31.00 ©2022 IEEE
}\\
Department of Electrical and Computer Engineering, Boston University\\
8 Saint Mary's Street, Boston, MA 02215\\
{\tt\small [mcokbas, jcbolo, jkonrad, pi]@bu.edu}
}

\maketitle
% \thispagestyle{empty}

%%%%%%%%% ABSTRACT
\begin{abstract}
Person re-identification (PRID) from side-mounted rectilinear-lens cameras is a well-studied problem. On the other hand, PRID from overhead fisheye cameras is new and largely unstudied, primarily due to the lack of suitable image datasets. To fill this void, we introduce the ``Fisheye Re-IDentification Dataset with Annotations'' (FRIDA)\footnote{\href{http://vip.bu.edu/frida}{\tt vip.bu.edu/frida} \\ 978-1-6654-6382-9/22/\$31.00 ©2022 IEEE}, with 240k+ bounding-box annotations of people, captured by 3 time-synchronized, ceiling-mounted fisheye cameras in a large indoor space. Due to a field-of-view overlap,
PRID in this case differs from a typical PRID problem, which we discuss in depth. We also evaluate the performance of 10 state-of-the-art PRID algorithms on FRIDA. We show that for 6 CNN-based algorithms, training on FRIDA boosts the performance by up to 11.64\% points in mAP compared to training on a common rectilinear-camera PRID dataset.
\vglue -0.5cm
\end{abstract}
% \let\thefootnote\relax\footnote{978-1-6654-6382-9/22/\$31.00 ©2022 IEEE}

%%%%%%%%% BODY TEXT
\section{Introduction}

Knowing the number and location of people in public spaces, office and school buildings, stores and shopping malls, etc., is critical for public safety (fire, chemical hazards), spatial analytics (optimization of office or store space usage), HVAC energy reduction, and, recently, for pandemic management. Typically, people-detection systems use standard surveillance cameras (equipped with rectilinear lens) mounted high on walls above the scene of interest. Since such cameras have a relatively narrow field of view (FOV), a number of them must be installed and managed which significantly increases the system complexity and cost, especially in large spaces.

Recently, overhead fisheye cameras have been successfully proposed for people counting \cite{Tamura, HABBOF, RAPiD}. However, even a fisheye camera with its large FOV cannot 
%capture sufficiently-fine images to 
accurately detect people at the FOV periphery (large distance from the camera) due to extreme foreshortening and geometric distortions. Clearly, in such spaces (e.g., a convention hall) multiple overhead fisheye cameras are needed. However, since the same person may appear in FOVs of multiple cameras, person re-identification (PRID) is critical for accurate people counting, tracking, etc.

%In recent years, due to the global pandemic, terms like ``room/building capacity'' have gained more significance in our daily lives. To automate the process of getting the people count of a room, researches have developed systems which use cameras to monitor the rooms. These systems use people detection algorithms to count the number of people. Most of these algorithms \cite{Tamura, HABBOF, RAPiD, WEPDTOF} were designed for images that are captured by overhead fisheye-lens cameras. The key advantage of the fisheye-lens cameras over the rectilinear-lens cameras is their wide field-of-view (FOV). Thus, a single fisheye camera can cover large spaces. The performance of the existing overhead fisheye people detection algorithms have been proven to be perform well in {\it small-to-medium} sized rooms. However, in large rooms, these methods start to struggle detecting people that are appearing at the periphery of the frame. This is due to the fact that as people get furhter from the fisheye camera they appear extremely small on the frame and get distorted by the fisheye lens.

%To monitor the people count in large rooms with overhead fisheye cameras, we can use multiple of them in a single room. However, this will arise the problem of double-counting of the people that are close to the both cameras. To overcome the double-counting problem, we can apply person re-identification (PRID) to identify the same people that are detected from the FOV of multiple cameras.

While PRID for {\it side-mounted, rectilinear-lens} cameras has been researched in depth 
%\cite{LF_ensemble, LF_salient_color, LF_symmetry, LF_and_metric,probabilistic_distance, KISS_distance, kernel_based_metric, relaxed_distance, PCB,ABD,SCAL, SONA, Pyramid, DL_harmonious, DL_unsupervised, DL_app_struc_gen, omni_feature, VA_ReID, CTL,PRID_overview}
\cite{PCB,ABD, Pyramid, VA_ReID, CTL,PRID_overview}, 
%the topic is barely explored for {\it overhead fisheye} cameras.
%\jlk{Do we need so many? We need papers describing the algorithms we are using, perhaps 2-3 "classics" and maybe a review article or two.}
%\mc{I picked 5 DL methods that we explore and a review article. Please let me know if this is still too much } 
we are aware of only three works exploring fisheye PRID \cite{fisheye_PRID_Barman, fisheye_PRID_Blott,  josh}, each with its own limitations and none releasing their fisheye data. Therefore, to inspire more research in this area, we are proposing a first-of-its-kind dataset, ``Fisheye Re-IDentification Dataset with Annotations'' (FRIDA), that was captured by three overhead fisheye cameras in a large space and includes over 240,000 bounding-box annotations of people.
%A comparison of FRIDA with the existing image-based PRID datasets can be seen in Table \ref{tab:Comparison of Typical Datasets}.
In addition to introducing FRIDA, we explore its use for image-based PRID. An alternative use-case for FRIDA is as a video-dataset for tracking, but this is not the focus of our work.

%FRIDA may be considered as either an image-dataset for PRID or as a video-dataset for tracking. In this paper we explore it in an image-based PRID context.

%Person re-identification with { \it side-mounted rectilinear-lens} cameras have been well-explored \cite{LF_ensemble, LF_salient_color, LF_symmetry, LF_and_metric,probabilistic_distance, KISS_distance, kernel_based_metric, relaxed_distance, PCB,ABD,SCAL, SONA, Pyramid, DL_harmonious, DL_unsupervised, DL_app_struc_gen, omni_feature, VA_ReID, CTL}. However, person re-identification in indoor spaces with an { \it overhead fisheye} cameras is an under-studied problem. There have been some attemps \cite{fisheye_PRID_Barman, fisheye_PRID_Blott,  josh} but these methods have some limitations and none of these studies published their data. To inspire more research in this area, we are a publishing a dataset that is first of its kind, called Fisheye Re-IDentification Dataset with Annotations (FRIDA). FRIDA is captured by three overhead fisheye cameras and has over 240,000 bounding boxes which are all manually labelled. A comparison of FRIDA with the existing image-based PRID datasets can be seen in Table \ref{tab:Comparison of Typical Datasets}. FRIDA can be treated as image-based or video-based PRID dataset, but in the scope of this paper, we will be treating it as an image-based PRID dataset due to the unique problem statement that we will be introducing in the upcoming paragraph.

Typical PRID datasets are not designed for people counting and were captured by side-mounted, rectilinear-lens cameras without FOV overlap. In this case, the goal is to identify the same person in two images captured by two cameras at {\it different} times. FRIDA, however, is meant for people counting and was captured by time-synchronized, overhead, fisheye cameras with {\it fully-overlapping} FOVs (360$^\circ\times$185$^\circ$).  In this case, the goal is to identify the same person in two images captured by two cameras at the {\it same} time. This explains the difference between the gallery sets of typical PRID datasets and FRIDA. In the former, for a given query there may be multiple ground-truth gallery elements, captured at different times. In FRIDA, for a given query there may be at most {\it one} gallery element at a given time instant. In case of occlusion, there is no gallery element for a given query (see Section~\ref{sec:Dataset} for details).

%This difference in the problem statement causes a key difference in between the sizes of the gallery sets of typical PRID datasets and FRIDA. In typical PRID datasets, there are multiple ground truth gallery elements for a given query which are captured at different time instants. On the other hand, in FRIDA, for a given query at a certain time instant, there can be at most one gallery element at a certain time instant. In the event of an occlusion, there can even be no gallery elements for a given query. In Section \ref{sec:Dataset}, we elaborate on the differences between typical PRID datasets and FRIDA in detail.

We also evaluate the performance of 10 state-of-the-art PRID methods on FRIDA: 6 methods developed for typical PRID datasets \cite{PCB, ABD, CTL, VA_ReID, Pyramid, ResNet} and 4 methods developed for overhead fisheye cameras \cite{josh}. The results show that training CNN-based methods on FRIDA (2-fold cross-validation) improves performance by 4.99-11.64\% points in mAP compared to training on a typical PRID dataset \cite{Market-1501}.

The main contributions of this work are:
\vspace{ -1ex}
\begin{itemize} \itemsep 0em
    \item We introduce a new PRID dataset, FRIDA, for indoor person re-identification using time-synchronized overhead fisheye cameras. This is the first overhead fisheye dataset for PRID and will be made publicly available.
    \item We evaluate the performance of 10 state-of-the-art PRID methods on FRIDA using two metrics. %\mcrep{We highlight the strengths and weaknesses of different algorithms.}{We provide comparison of these algorithms with respect to each other.} \jlk{Do we highlight?} \mc{I think what I was doing was not highlighting. So, I re-phrased it}
    We compare the performance of 6 of those algorithms, when training on FRIDA against training on the non-fisheye Market-1501 dataset \cite{Market-1501}.
\end{itemize}

% As a motivation, mention the people counting application. Large room -> hard to cover -> Fisheye Cameras -> Distortion -> Multiple Cameras -> Double Count -> PRID.

% PRID with fisheye cameras is an under-studied problem. We think the main reason for this is the lack of a dataset, we are hoping FRIDA will inspire some new research in this area.

\vspace{ -2.5ex}
%-------------------------------------------------------------------------------
\section{Related Work}
\subsection{Datasets}
\vglue -0.2cm
There exist several datasets for person re-identification using side-mounted rectilinear-lens cameras. Table~\ref{tab:Comparison of Typical Datasets} lists key statistics of the most common ones: VIPeR \cite{VIPeR}, PRID 2011 \cite{PRID_2011}, Airport \cite{Airport}, CUHK03 \cite{CUHK03}, GRID \cite{GRID},  MSMT17 \cite{MSMT17}, Market-1501 \cite{Market-1501} and iLIDS \cite{iLIDS}, but more details can be found in \cite{PRID_overview}. All these datasets have been designed with the goal of matching the image of a person from the query set to an image from the gallery set, and the query and gallery sets consist of images captured by {\it different} cameras. Moreover, different cameras have no field-of-view overlap so query and gallery images of the same identity have been captured at {\it different time instants}.
%
%\picomment{Are you sure there is no FOV overlap? Even if there is no FOV overlap how can you be sure that they were captured at different instances? Are these explicitly asserted in the cited references, or are you guessing? }
%
Finally, in most of these datasets there are, typically, \textit{multiple} gallery images having the same ID as the query image.

While there exist people-focused datasets captured by overhead fisheye-lens cameras (PIROPO \cite{PIROPO}, BOMNI \cite{BOMNI}, MW \cite{MW}, HABBOF \cite{HABBOF}, CEPDOF \cite{RAPiD}, WEPDTOF \cite{WEPDTOF}), 
%\jlk{This is too VIP focused. I would refer to MW instead of MW-R, but I am not sure MW has ID. How about PIROPO and others?} \mc{I added PIROPO, BOMNI and changed MW-R to MW}
they have been developed with the goal of people detection and, in some cases, tracking. However, each dataset only consists of frames from a \textit{single} camera which severely limits the variability of body appearance, unlike in FRIDA.
%and thus none of them is suitable for cross-camera person re-identification (although they could potentially be used for cross-time PRID from the same camera). 

\begin{table}[!htb]
\caption{Commonly-used image datasets for person re-identification. (BBox = bounding box)}
   \label{tab:Comparison of Typical Datasets}
\centering
\small
\smallskip
\begin{tabular}{|c|c|c|c|c|}
   \hline
 Dataset & Year & \# & \# & Frame \\
 & & BBoxes & Cameras & Resol.\\
 \hline
VIPer \cite{VIPeR} & 2007 & 1,264 & 2 & Fixed \\
\hline
iLIDS \cite{iLIDS} & 2009 & 476 & 2 & Variable \\
\hline
GRID \cite{GRID}& 2009 & 1,275 & 8 & Variable \\
\hline
PRID 2011 \cite{PRID_2011}& 2011 & 24,541 & 2 & Fixed \\
\hline
CUHK03 \cite{CUHK03}& 2014 & 13,164 & 2 &  Variable \\
\hline
Market-1501 \cite{Market-1501}& 2015 & 32,668 & 6 &  Fixed\\
\hline
Airport \cite{Airport}& 2017 & 39,902 & 6 & Fixed \\
\hline
MSMT17 \cite{MSMT17}& 2018 & 126,441 & 15 &  Variable\\
\hline
{ \bf FRIDA} & 2022 & 242,809 & 3 & Fixed \\
\hline
\end{tabular}
\vglue -0.4cm
\end{table}

\subsection{Algorithms}

Person re-identification using rectilinear-lens cameras is a well-studied problem. Early approaches were model-based \cite{LF_ensemble, LF_symmetry, LF_and_metric, KISS_distance} and used hand-crafted features. Recent approaches use deep learning \cite{PCB,ABD,SCAL, SONA, Pyramid, VA_ReID, CTL, PRID_overview} and outperform the traditional methods.
%We review key methods below.
%\jlk{I would prune the number of references if we need space. Usually, referring to a good review article solves the problem.} \mc{I pruned it, please let me know if I should prune further or not.}

%Tradiitonal ref before pruning \cite{LF_ensemble, LF_salient_color, LF_symmetry, LF_and_metric,probabilistic_distance, KISS_distance, kernel_based_metric, relaxed_distance}
% DL refs before pruning \cite{PCB,ABD,SCAL, SONA, Pyramid, DL_harmonious, DL_unsupervised, DL_app_struc_gen, omni_feature, VA_ReID, CTL} 

Sun {\it et al.}~proposed PCB \cite{PCB} in which feature vectors are uniformly partitioned in an intermediate layer to obtain part-informed features. This structure allows to separately focus on different parts of an image and extract local information for each part.  Zheng {\it et al.}~proposed a network called Pyramid \cite{Pyramid} which does not only focus on part-informed local features, but also accounts for global features in addition to gradual cues. Pyramid achieves this through a coarse-to-fine model, which performs image matching by leveraging information from different spatial scales. Chen {\it et al.}~proposed an attention-based network called ABD-Net \cite{ABD}, which instead of a small portion of an image focuses on its wider aspect by means of a diverse attention map. This is accomplished by combining two separate modules: one module focuses on context-wise relevance of pixels while the other module focuses on spatial relevance of these pixels. Zhu {\it et al.}~proposed a network called VA-reID \cite{VA_ReID} that allows matching of people regardless of the viewpoint from which they were captured. Instead of creating a separate space for each viewpoint (i.e., front, side, back), they create a unified hyperspace which accommodates viewpoints in-between the main viewpoints (e.g, side-front, side-back, etc.). Recently, Wieczorek {\it et al.}~proposed a CTL model (Centroid Triplet Loss model) \cite{CTL}, which extends the triplet loss. When working with triplet loss, it is typical to choose one positive sample and one negative sample for an anchor. However, in the CTL model, instead of choosing a single sample, a centroid is computed over a set of samples which significantly improves performance.

The methods above have been designed for and tested on images from rectilinear-lens cameras. Very few PRID methods have been developed for overhead fisheye cameras. An early approach, proposed by Barman {\it et al.}~\cite{fisheye_PRID_Barman}, matches images of people who appear at the same radial distance from a camera (similar viewpoints). This is restrictive, and leads to sub-par performance, since people often appear at different distances from FOV centers in different cameras. Another algorithm proposed by Blott {\it et al.}~\cite{fisheye_PRID_Blott} applies tracking to extract front-, back- and side-view images of a person. A person-descriptor is built by fusing features extracted from individual views. The algorithm does not perform PRID for each pose/viewpoint. Moreover, there is no guarantee that a person will appear at all 3 viewpoints during a recording, thus limiting performance.
%
%Both of these methods have limited applicability since they assume an upright standing position or require orientation normalization, but in indoor scenarios people often sit, bend over a laptop, lean on a table, etc. 
%
Recently, Bone {\it et al.}~\cite{josh} proposed a PRID method for fisheye-lens cameras with overlapping FOVs. This approach leverages locations of people in images {\it instead} of their appearance. Using a calibrated fisheye-lens model this method maps pixel-location of a person in a query image to a pixel-location in a gallery image. The mapped query-person location is compared to locations of people in the gallery image to establish a match. The advantage of this algorithm is that it does not rely on a person's appearance, but it requires camera calibration (intrinsic parameters) and the knowledge of camera position in the 3D world (distance/rotation between cameras, mounting height). No datasets were published from these studies.
%
%The datasets from neither of these three fisheye PRID studies is publicly available.

% General PRID

% Fisheye PRID (that has not datasets)

% Geometric PRID

\begin{table*}[htb]
\caption{Detailed information about FRIDA (Fisheye Re-Identification Dataset with Annotations).}
   \label{tab:Dataset}
\centering
\smallskip
\begin{tabularx}{\textwidth} { 
  | >{\hsize=0.2\hsize\centering\arraybackslash}X 
  | >{\hsize=0.16\hsize\centering\arraybackslash}X |
   >{\hsize=0.21\hsize\centering\arraybackslash}X |
   >{\hsize=0.18\hsize\centering\arraybackslash}X |
   >{\hsize=1.25\hsize\arraybackslash}X |}
  
 \hline
  & \# frames & \# BBoxes & \# BBoxes per frame & Scenarios/Challenges\\
\hline
Segment 1  & 7,017 & 66,810 & 3-15 & People coming in and settling down; evenly distributed around the room; mostly sitting (lower bodies mostly occluded)\\
\hline
Segment 2  & 3,471 & 53,460 & 13-18 & People walking around the room; significant occlusions\\
\hline
Segment 3 &  6,207  & 103,141 & 13-17 & Concentration of people in parts of the room; people standing and staying close to each other; people strongly occluding each other\\
\hline
Segment 4 &  1,623  & 20,028 & 5-16 &  People leaving the room; occasional occlusions at entry/exit points\\
\hline
\end{tabularx}
\vskip -0.3cm
\vglue -0.1cm
\end{table*}
%-------------------------------------------------------------------------------
\vspace{-2ex}
\section{FRIDA Dataset} \label{sec:Dataset}

%The development of FRIDA was driven by the need to efficiently and effectively monitor occupancy in large indoor spaces. Owing to their large field of view (180$^\circ\ \times$ 360$^\circ$), ceiling-mounted fisheye cameras are often the sensing modality of choice. Since a single fisheye camera is not fully effective in a space of thousands of square feet, multiple such cameras are needed thus calling for PRID. 

\begin{figure}[t]
\centering
\vglue -0.4cm
   \includegraphics[width=0.6\linewidth]{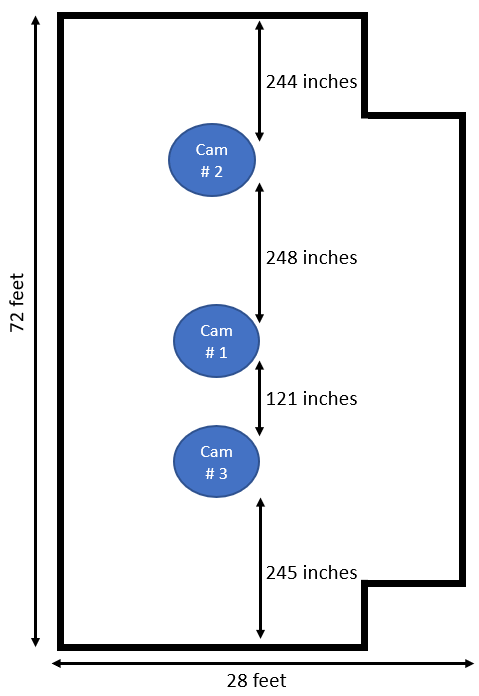}
   \caption{Bird's eye view of the space where FRIDA was collected.}
   \label{fig:camera_layout}
   \vglue -0.4cm
\end{figure}

\begin{figure}[p]
  \centering
  \includegraphics[width=0.8\linewidth]{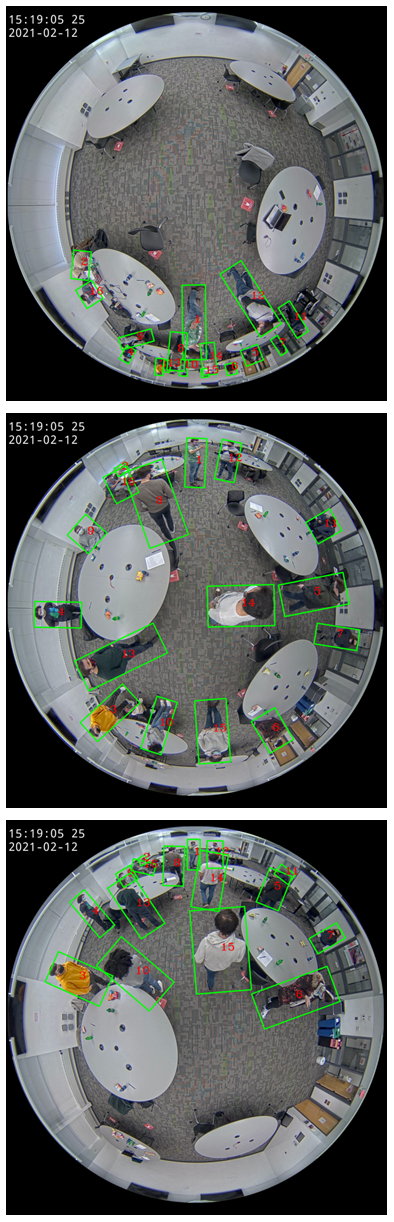}
  \caption{Example of three synchronously-captured fisheye images with annotations from FRIDA (top: camera 2, middle: camera 1, bottom: camera 3). 
%  \picomment{To help the reader easily locate BBs of person \#14 and \#15, their BB's should be highlighted using distinctive colors.}
  \label{fig:sample_triplet}
  }
\end{figure}

FRIDA is the first PRID dataset captured indoors by multiple overhead fisheye cameras and will be made publicly available\footnote{\href{http://vip.bu.edu/frida}{\tt vip.bu.edu/frida}}.
In FRIDA, the cameras have fully-overlapping FOVs (360$^\circ\times$ 185$^\circ$), unlike in typical PRID datasets, and are time-synchronized (frames are captured at the same time). FRIDA was collected in a 2,000ft$^2$ room using 3 ceiling-mounted fisheye cameras (100in above the ground). The bird's eye view of the room is shown in Fig.~\ref{fig:camera_layout}, and an example of time-synchronized frame triplet is shown in Fig.~\ref{fig:sample_triplet}.  The frames were captured by three Axis M3057-PLVE cameras at 2,048$\times$2,048-pixel resolution and 1.5 frames/sec. Annotations in FRIDA consist of 242,809 bounding boxes drawn around people (Table~\ref{tab:Comparison of Typical Datasets}).

FRIDA can be used in a number of ways: as an image-based dataset for PRID, as a video-based dataset for people-tracking, or just for people detection and counting.
In this paper, to demonstrate its most unique features, we treat it as an image-based PRID dataset. Below, we discuss the unique characteristics and challenges of FRIDA.

%{\bf Problem Statement:} The main goal in FRIDA is to perform person re-identification between fisheye frame pairs that are captured at the {\it same time instant} by overhead fisheye cameras which have fully overlapping FOV . 

{\bf Annotations:} At each time instant, three video frames are available with manually drawn, human-aligned bounding boxes for all people visible in each frame. Each bounding box is represented by 6 parameters: $x,y,w,h,\alpha, ID$, where $(x,y)$ are the coordinates of its center, $(w,h)$ are its width and height, $\alpha$ is its counter-clockwise rotation angle with respect to the vertical axis of the image,
and $ID$ is the ID number of a person. Each person in the dataset is assigned a unique ID which is consistent in all frames of the dataset. There are 20 unique ID numbers in FRIDA.

{\bf Scenarios:} FRIDA consists of four segments where each segment captures a different type of challenge (Table \ref{tab:Dataset}). In segment \#1, people enter the room, walk and sit down (people are evenly distributed in the room). This segment resembles a lecture where people remain seated for most of the time and their lower bodies are mostly occluded. Segment \#2 is the most crowded and dynamic segment. People are constantly moving which occasionally causes severe occlusions, especially when people are close to each other. This segment resembles a social meeting where people are wandering around the room and talking to each other. Segment \#3 is the longest  one and has over 100,000 bounding boxes. Participants gather at either end or in the middle of the room. They stand close to each other leading to severe occlusions. Segment \#4 is the shortest, with people leaving the room and causing occasional occlusions at the doors.

%{\bf Number of Ground Truth in the Gallery Sets:}
{\bf Single sample of ID in the gallery set:} In typical PRID datasets, for a given query element there are multiple samples in the gallery set with the query ID. In FRIDA, however, frames are captured at the same time instant and the identities in one frame are treated as the query set while identities in another frame are treated as the gallery set. Therefore, for a given query element there can be at most one sample with the query identity in the gallery set. In some cases, due to occlusions, a person may not be visible from a camera. This may lead to a no-match scenario at certain time instants for some query elements. Note that FRIDA can also be used for typical PRID by constructing the gallery from multiple images of the same ID captured at different times, but this is not in the scope of this work.

{\bf Synchronous, overhead capture:} Due to the overhead placement of cameras and simultaneous capture, the viewpoint of a person directly under one camera may be dramatically different from the viewpoint from another camera. This is unlike in most other PRID datasets where it is common to capture a person from similar viewing angles (e.g., front, back, side, top) using different cameras. Then, if one of the gallery elements has the same viewpoint as the query, the chance of a match increases. However, in FRIDA, since the query and gallery elements are synchronously recorded by different overhead cameras, people never appear from the same viewpoint. This can be seen in Fig.~\ref{fig:sample_triplet} where person \#14 is seen from the top in camera \#1 view, from the front in camera \#3 view and from the back in camera \#2 view. This makes the problem of PRID more challenging in FRIDA compared to other datasets.

%\jbrep{Especially, if a person is far from the camera in 3D-world (i.e. appearing around the periphery of the frame), due to the geometry of the fisheye-lens the distortions get severe.}{The distortions get especially severe if a person is particularly far from the camera in the 3D world (i.e. appearing around the periphery of the frame) due to the geometry of the fisheye lens.}

%{\bf The Viewing Angle Difference between the Query and the Gallery:} In the typical PRID datasets, it is common to capture the same person from similar viewing angle (i.e. front, back, side, top) from different cameras. In these datasets, for a query element, if one of the ground truths, out of the many, is captured from a similar angle and view, then, the possibility of having a correct match increases. However, in FRIDA, since the query and gallery elements are captured by different cameras at the same time instant, the majority of the time people are captured from different viewing angles . An example of this can be seen in Fig \ref{fig:sample_triplet}, person \# 14 is captured from top in camera \#1 view, from front in camera \#3 view and from back in camera \#2 view.

{\bf Fisheye distortions:} Since FRIDA has been recorded by fisheye cameras, images are subject to radial geometric distortions, especially close to FOV periphery. When a person is located at a different distance to each camera, the person's appearance is geometrically distorted to a different degree in each camera view. This makes the problem of PRID more challenging in FRIDA compared to other datasets.

\begin{figure}[b]
\centering
\vglue -0.4cm
   \includegraphics[width=0.95\linewidth]{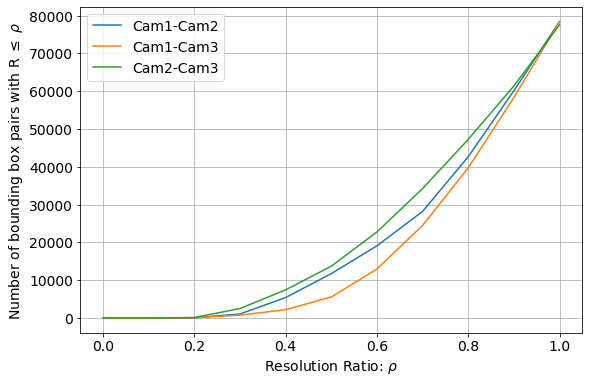}
   \caption{Bounding-box resolution mismatch for all camera pairs. 
%   \picomment{The horizontal axis should be labeled `Resolution Ratio: $\rho$' and the vertical axis should be labeled `Number of bounding-pairs with $R \leq \rho$'.}
   \label{fig:resolution_mismatch}}
\end{figure}

{\bf Resolution mismatch between query and gallery sets:} The synchronous, overhead capture and fisheye distortions often lead to very differently-sized bounding boxes for the same person (resolution mismatch). 
%
%at the same time instant from different camera views, they often happened to be closer to a certain camera compared to the other two cameras. This creates resolution difference between query and gallery elements.
%
Examples can be seen in Fig.~\ref{fig:sample_triplet}, e.g., person \#15 appears with very different resolutions in camera \#2 and camera \#3 views. In Fig.~\ref{fig:resolution_mismatch}, we demonstrate this resolution mismatch quantitatively. The resolution ratio $R$ between two bounding boxes $B_1$ and $B_2$ is defined as follows:
\vspace{-2ex}
\begin{equation} \label{eqn:Resolution_ratio}
  R = \frac{min(Area(B_1), Area(B_2))}{max(Area(B_1), Area(B_2))}.
  \vspace{-2ex}
\end{equation}
Each data point in the plot shows the number of bounding-box pairs such that $R\leq \rho$ with $0\leq\rho\leq 1$. Note that, the resolution mismatch is the largest (highest curve) between cameras 2 and 3 since they are farthest apart (Fig.~\ref{fig:camera_layout}).

% TO DO:

% \mcrep{}{Table with other datasets (camera count, number of bounding boxes, total images)}

%-------------------------------------------------------------------------------
\section{Experiments}

\subsection{Algorithms}

In order to gauge challenges offered by FRIDA, we evaluated ten state-of-the-art PRID algorithms on it. Six of these algorithms: PCB \cite{PCB}, Pyramid \cite{Pyramid}, ABD-Net \cite{ABD}, VA-reID \cite{VA_ReID}, CTL \cite{CTL} and ResNet-50 \cite{ResNet}, are CNN-based and use person's appearance for re-identification. The remaining four algorithms are geometry-based and use person's location in fisheye images for re-identification \cite{josh}. They  differ in terms of the re-identification metric used: point-to-point distance (PPD), point-to-set total distance (PSTD), point-to-set minimum distance (PSMD) and count-based distance (CBD). FRIDA's annotations serve as the input for all methods: bounding-box RGB values for appearance-based methods and bounding-box center for geometry-based methods. In each algorithm, we used hyper-parameters suggested in the corresponding paper.

%People detection is out of the scope of this paper. Thus, during our experiments, for people detections we used the ground truth bounding boxes that are given by the manual annotations of FRIDA.
\begin{table*}[htb]
\caption{Performance comparison of the appearance-based algorithms trained on {\bf Market-1501} and tested on {\bf FRIDA}. The highest values of QMS and mAP for each camera pair and for the cumulative are shown in boldface.}
\label{tab:Market_1501_res_cross_validation}
\smallskip
\centering
\begin{tabular}{|c|c|c|c|c|c|c|c|c|}
\hline
&
\multicolumn{4}{c|}{QMS [\%]} &
\multicolumn{4}{c|}{mAP [\%]} \\
\hline
& C.1+C.2 & C.1+C.3 & C.2+C.3 & Cumulative &
C.1+C.2 & C.1+C.3 & C.2+C.3 & Cumulative \\
\hline
ResNet-50 \cite{ResNet} & 57.63 & 73.99 & 45.33 & 59.04 & 70.03 & 79.41 & 59.33 & 69.63\\
\hline
PCB \cite{PCB} & 56.63 & 74.64 & 45.62 & 59.02 & 70.91 & 79.28 & 59.79 & 70.04\\
\hline
ABD \cite{ABD} & 61.26  & 73.68 & 44.22 & 59.78 & 70.80 & 77.93 & 58.71 & 69.19\\
\hline
Pyramid \cite{Pyramid} & {\bf 74.58} & {\bf 84.66} & {\bf 54.88} & {\bf 71.46} & {\bf 78.72} & {\bf 86.38} & {\bf 64.89} & {\bf 76.72} \\
\hline
VA-ReID \cite{VA_ReID} & 60.79 & 74.18 & 44.99 & 60.06 & 71.21 & 78.68 & 59.31 & 69.78 \\
\hline
CTL \cite{CTL} & 66.92 & 83.68 & 42.88 & 64.59 & 72.57 & 84.99 & 57.44 & 71.72\\
\hline

\end{tabular}
\end{table*}

\begin{table*}[htb]
\caption{Performance comparison of the appearance-based algorithms trained on {\bf FRIDA} and tested on {\bf FRIDA}. The highest values for QMS and mAP for each camera pair and for the cumulative are shown in boldface.}
\label{tab:FRIDA_res_cross_validation}
\smallskip
\centering
\begin{tabular}{|c|c|c|c|c|c|c|c|c|}
\hline
&
\multicolumn{4}{c|}{QMS [\%]} &
\multicolumn{4}{c|}{mAP [\%]} \\
\hline
& C.1+C.2 & C.1+C.3 & C.2+C.3 & Cumulative &
C.1+C.2 & C.1+C.3 & C.2+C.3 & Cumulative \\
\hline
ResNet-50 \cite{ResNet} & 64.93 & 75.79 & 50.11 & 63.67 & 76.20 & 81.60 & 68.00 & 75.30 \\
\hline
PCB \cite{PCB} & 63.30 & 74.79 & 51.77 & 63.33 & 75.79 & 81.23 & 67.91 & 75.01\\
\hline
ABD \cite{ABD} & 75.31 & 83.18 & 62.05 & 73.57 & 82.43 & 85.16 & 74.81 & 80.83\\
\hline
Pyramid \cite{Pyramid} & 67.79 & 80.78 & 53.48 & 67.42 & 75.38 & 81.59 & 68.61 & 75.23  \\
\hline
VA-ReID \cite{VA_ReID} & 67.52 & 79.46 & 54.59 & 67.24 & 76.58 & 82.74 & 68.00 & 75.81 \\
\hline
CTL \cite{CTL} & {\bf 77.30} & {\bf 90.11} & {\bf 64.76} & {\bf 77.44} & {\bf 82.7} & {\bf 89.79} & {\bf 75.17} & {\bf 82.58}\\
\hline
\end{tabular}
\vglue -0.4cm
\end{table*}
\vspace{-2ex}
\subsection{Implementation Details}

%For CNN-based algorithms, we trained them separately on two different datasets. The first dataset that we trained them on is \jbedit{the commonly-used} Market-1501 \cite{Market-1501}, \jbedit{commonly used in rectilinear PRID.} \jbrep{and t}{T}he second dataset we used to train the networks is WEPDTOF \cite{WEPDTOF}. We wanted to demonstrate the performance difference between training the networks on \jbedit{a} side-view rectilinear-lens camera dataset with high number of identities and training the networks on a overhead-fisheye dataset with relatively less number of identities. While using WEPDTOF for training, we zero-padded and resized the bounding boxes to match the input size of the networks.  We kept whole FRIDA as \jbedit{the} testing set (i.e. did not use it in training). \mc{Discuss with John, the optimizer used, the learning rate, the number of iterations, momentum and weight decay. Check whether it is the same for all networks, if not , we might need to do a table for all these parameters}

The testing procedure for all methods was the same. For a given pair of video frames, we treated all people from one frame as as the query set (images or locations) and those from the other frame as the gallery set (images or locations).

In CNN-based methods, we fed the image of a person (within the person's bounding box) into a network and extracted a feature vector from the final convolutional layer to serve as this image's descriptor.  We computed the cosine similarity between all normalized feature vectors of the query and gallery sets, resulting in a score matrix for each pair of frames\footnote{Since cosine distance is symmetric, it is not important which frame is chosen as the query set and which as the gallery set.}. We applied greedy matching to the score matrix to match the query and gallery elements as follows. We found the highest matching score, considered the corresponding elements to be a match and removed their row and column from the table. We repeated this process until no more removals were possible. We used this algorithm since in FRIDA the fisheye cameras have a fully-overlapping FOV and, therefore, a person can have at most {\it one} match in another camera's FOV (and can be removed from the score matrix).
%
%Reason for using greedy algorithm is the fact that cameras have fully overlapping FOVs. In other words, since the cameras have overlapping FOV, the person is going to have at most one single match, so, if a person is matched, then s/he can be removed from the score matrix.
%
This is different than testing on typical PRID datasets, where the matching of a given query does not affect the matching of other query elements.
%However, in our case, since we apply greedy matching, the matching of a query element is dependent on the other query elements.

%\jlk{I think all this is unclear. First, in Market-1501 there are no frames, right? There are just person-images (bounding boxes) that are not associated with frames. One simply forms query and gallery sets from those images, right? This is different in WEPDTOF. In fact, removed any reference to camera views since in WEPDTOF there is only one camera per sequence. Different people are in each sequence (except, potentially, for the split sequences).}
%\mc{The paragraph above describes the testing procedure and it is mentioned that for testing we just used FRIDA. So, not sure if I understood your question. I would be happy to discuss this in our meeting.}
%\jlk{Secondly, is the meaning of "greedy matching" obvious? That is that the top match is removed and then the process repeated? I believe our approach to matching is due to the fact that we have overlapping FOVs so ther is at most a single match, unlike in typical PRID where multiple matches can take place. I think this paragraph needs to be carefully re-written.}
%\mc{You are right, I agree that I did not explain it why we use greedy algorithm very clearly. I made some adjustments to address this issue.}

In geometry-based algorithms, we first performed camera calibration as outlined in \cite{josh}. Using the calibrated camera model, each algorithm maps pixel locations of people (centers of bounding boxes) in the query frame (set) to pixel locations in the gallery frame (set). The matching is performed using different distance measures between the mapped query locations and gallery locations.
%assuming either a 1-to-1 match or no match occurs.
Importantly, the geometric mapping depends on the height of a person \cite{josh}. In PPD, the mapping is performed for an average person's height (168cm) and the Euclidean distance from the mapped query location to a gallery location serves as the distance measure. In PSTD, query locations are mapped to the gallery frame for 21 different heights of a person (128-208cm). The {\it sum} of the Euclidean distances from 21 mapped query locations to a gallery location serves as the distance measure. The PSMD algorithm is very similar except that the {\it minimum} of 21 Euclidean distances instead of their sum is used as the distance measure. In CBD, again 21 mapped query locations are produced and the number of such locations that are closest to a gallery location is used to compute a distance measure. The distance measure computed in each case is used to construct a distance matrix for greedy matching as described above (except that the {\it smallest} distance is considered a match). Since so-defined distance measures are not symmetric, we performed bidirectional matching. For more details, see \cite{josh}. 

\subsection{Dataset Splits}
\vglue -0.1cm
%\mcrep{We separately trained the CNN-based algorithms on Market-1501 \cite{Market-1501} and WEPDTOF \cite{WEPDTOF} datasets. Market-1501 is a commonly-used PRID dataset composed of images captured by side-view, rectilinear-lens cameras (different cameras capture a person at different times). WEPDTOF is a person detection and tracking dataset composed of sequences of video frames each captured by a different overhead fisheye camera (each person appears in view of a single camera only). In this experiment, we want to demonstrate a performance gap between training the networks on a standard-image dataset with a high number of identities and a fisheye-image dataset with relatively few identities. When training on WEPDTOF, we zero-padded/re-sized the bounding boxes to match the input size for each network.  We used FRIDA only for testing each algorithm.}{}

Despite more than 240,000 bounding boxes, FRIDA has only 20 different identities, Since this is insufficient for separate training, validation and testing sets, we evaluate the algorithms using 2-fold \textit{identity-wise} cross-validation. We use half of the identities in training and the other half in testing, and then we swap the roles of identities and repeat the process. Specifically, we created the training set by choosing 50 random time stamps for each identity and taking 3 images (one from each camera) captured at this time (cameras are synchronized). This allowed a rich training set with many different viewpoints of the same person. 
%
%Note that we removed images where a person was wearing different clothing than during the majority of the recording. This is important, since during testing on FRIDA PRID is performed between frames captured at the same time so a person cannot appear in different outfits in different camera views. 
%
%\picomment{If that is the case, then why there a need to remove? The removal may result an unequal number of examples for each identity. A simpler thing to have done is to just choose 50 examples for each identity randomly from the frames where the person has the same outfit as in the majority of time instants of the recording.}
%
We will provide the training sets we used as part of FRIDA.
%
%\picomment{Since examples are chosen at random time instants, I wonder in how many of the sampled instants will we have people wearing an outfit different from that in the majority of the recording. Is it really that significant a number? Is it really worth introducing this manual post-processing? In any case, we need to carefully rephrase the last 2 sentences.}
%
%\jlk{I thought FRIDA was recorded in one sitting and people did not change clothing. Did people remove jackets? Still, this is the same person. Is it assumed in typical PRID datasets that people have the same clothing?} 
%
%\mc{It was winter and people came with their coats to the recording. Once they sit down, they took off their coats. Some people did not take their coats at all. In typical PRID it would be considered the same person, although in Market-1501 to the best of my knowledge there is no cloth change. In our test setup there is no way, a person will appear in different clothing at the same time instant in different camera views, so that's why we pruned those bounding boxes for training set.}
%
%\jlk{Will the information about clothing changes be provided in the published dataset so that others can use it?} 
%
%\mc{I think this is a good idea for reproducibility. We can provide the training sets as separate folders in the dataset.}
%
In the testing sets, we use {\it all} frames with the specific identities. 

We also trained the CNN-based networks on Market-1501 \cite{Market-1501} and tested them on FRIDA. Market-1501 is a commonly-used PRID dataset composed of images captured by side-view, rectilinear-lens cameras (different cameras capture a person at different times). For fairness, we used the same cross-validation testing sets as when both training and testing on FRIDA.
%and we averaged the results.
We used the same testing sets for the geometry-based algorithms. 

%Then, again, we tested these networks on FRIDA with the same set of cross-validation identities. Note that, in this case, training is on purely Market-1501, but for fair comparison, the testing is performed in 2-splits. We tested the geometry-based algorithms again in the same fashion where we used the same 2-fold cross-validation splits.

\vspace{-1ex}
\subsection{Evaluation Metrics}
\vglue -0.1cm
We use the {\it Query Matching Score} (QMS) \cite{josh} and {\it mean Average Precision} (mAP) as evaluation metrics. QMS is very similar to the commonly-used {\it Correct Matching Score} (CMS), and is defined as follows:
\begin{equation*}
  QMS = \frac{\sum_{n=1}^{N} \sum_{q\in Q_n} \mathbbm{1}(q=\widehat{q})}{\sum_{n=1}^{N} \big|Q_n \cap G_n\big|}
\end{equation*}
where $N$ denotes the number of frames, $Q_n, G_n$ are the sets of query and gallery identities in frame number $n$, respectively, and $\widehat{q}$ is the predicted identity of query $q$ or ``null'' if there is no match. The important difference between QMS and CMS is that QMS accounts for situations when there is no match between a query and gallery elements ($|Q_n\cap G_n|$ in the denominator). Basically, QMS gives the ratio of the number of correct matches to the number of true matches.
%
%\jlk{I think that perhaps we should explicitly define QMS since this is not a well-known metric. BTW, does CMS assume that there is always a match?}
%\mc{I put the definition of QMS above. I took it from Josh's paper and tried to rephrase it to my best ability. But if it is still too similar, I would appreciate some help to make it look more different. Yes, CMS assumes there is always a match}
%
We also compute the commonly-used mAP. It is important to note that in our scenario there exists {\it at most} one matching gallery-frame identity for a given query element.
Unlike in classical PRID, we can encounter a query whose identity is absent from the gallery (due to complete occlusion). We exclude such cases from the mAP calculation.
%In case a person is absent from the gallery (due to complete occlusion), a query element will not contribute to the mAP.
%
%\jlk{Meaning that an occlusion will not lower mAP?}
%
%\mc{I tried to mean full occlusion so the person is not visible at all and not labelled with a bounding box. On the other hand, there are partial occlusions too, and in those cases the person has a bounding box and contributes to the mAP. Mostly, partial occlusions make PRID difficult and lower the mAP.}
%
%\jlk{You mean that if a person is visible in one camera but fully occluded in another camera, there will be no match (the person in the first camera is not matched to anybody), right? And this will not affect mAP, right? What happens in QMS? If the first-camera identity is left out in the score matrix at the end, this ID has no match. How is this handled in the QMS formula?} 
%
%\mc{It is handled by the denominator of QMS. Because we are taking the cardinalty of the intersection of gallery set and query set. So, if someone has not match they will not be contributing to QMS. I added a colored text above to clarify it, please feel free to change it}

\begin{table*}[!htb]
\caption{Performance comparison of the geometry-based algorithms \cite{josh} on {\bf FRIDA}. The highest values for QMS and mAP for each algorithm are shown in boldface.}
\label{tab:geo_segment_results_cross_validation}
\smallskip
\centering
\begin{tabular}{|c|c|c|c|c|c|c|c|c|c|c|}
\hline
&
\multicolumn{5}{c|}{QMS [\%]} &
\multicolumn{5}{c|}{mAP [\%]} \\
\hline
& Seg.1 & Seg.2 & Seg.3 & Seg.4 & Cumulative &
Seg.1 & Seg.2 & Seg.3 & Seg.4 & Cumulative\\
\hline
PPD & 99.51 & 87.39 & 80.60 & 90.77 & 88.02 & 99.49 & 94.47 & 90.13 & 93.95 & 93.93\\
\hline
PSMD & 99.58 & 91.21 & 85.69 & 87.99 & 90.75 & {\bf 99.83} & 95.62 & 91.98 & 94.16 & 95.06\\
\hline
PSTD & {\bf 99.69} & 88.08 & 81.79 & 91.03 & 88.76 & 99.60 & 94.69 & 90.22 & 94.19 & 94.06\\
\hline
CBD & 99.17 & {\bf 92.18} & {\bf 89.81} & {\bf 93.69} & {\bf 93.11} & 99.69 & {\bf 96.82} & {\bf 95.41} & {\bf 96.51} & {\bf 96.97} \\
\hline
\end{tabular}
%\vglue -0.3cm
\vglue -3ex
\end{table*}
\vspace{-2ex}
\subsection{Results}

In Table \ref{tab:Market_1501_res_cross_validation}, we report results for the six appearance-based CNN algorithms trained on Market-1501 and tested on FRIDA. In Table \ref{tab:FRIDA_res_cross_validation}, we report results for the same algorithms, but both trained and tested on FRIDA. These results are computed over all 4 segments of FRIDA for each camera pair. 
%
%\mcrep{}
We also report the cumulative QMS value which is computed as the total number of correct matches from all camera pairs and all segments divided by the total number of possible correct matches from all camera pairs and all segments. In addition to QMS, we report mAP (the cumulative mAP is computed in a manner analogous to cumulative QMS).
The common trend in both tables is that all algorithms achieve the highest QMS/mAP for cameras 1 and 3, and the lowest for cameras 2 and 3. This was to be expected since cameras 1 and 3 are the closest to each other (Fig.~\ref{fig:camera_layout}); people are captured at a more similar resolution, viewpoint and geometric distortion compared to other camera pairs. Conversely, the distance between cameras 2 and 3 is the largest which makes PRID more challenging. 
%\jlk{Shouldn't camera distance also pose challenge to geometric PRID?} 
%\mc{You are right. Geometry-based algorithms also perform the worst on cam2 to cam3 pair. But the performance drop is not as high as appearance-based algorithms.}

As Table \ref{tab:Market_1501_res_cross_validation} shows, when trained on Market-1501, Pyramid \cite{Pyramid} performs the best among the six appearance-based methods and outperforms the second-best algorithm, CTL \cite{CTL},  by 6.87\% points in terms of cumulative QMS, and by 5.0\% points in terms of cumulative mAP. 
%
%\piedit{Note, that pyramid achieves an mAP of 86.38\% for cameras 1 and 3.}{} 
%
%\picomment{What is the point of noting the result for Pyramid for cameras 1 and 3? This is unclear and it seems not needed.}
%
% \picomment{Search for occurrences of mAP that need to be changed to cumulative mAP,}

When these algorithms are trained on FRIDA (Table \ref{tab:FRIDA_res_cross_validation}), CTL \cite{CTL} outperforms other networks by 3.87-14.11\% points in cumulative QMS and by 1.75-7.57\% points in cumulative mAP. For cameras 1 and 3, CTL performs above 90\% in terms of QMS. When trained on FRIDA, all networks achieve cumulative QMS above 63\% and cumulative mAP above 75\%.
%
%\jlk{I do not think it is true, e.g., ResNet50.} \mc{I was mentioning average QMS and mAP, here}

Comparing the performance of algorithms trained on Market-1501 versus those trained on FRIDA, all the networks performed better when trained on FRIDA except for Pyramid. In terms of cumulative QMS, the improvement achieved by training ResNet-50, PCB, ABD, VA-ReID and CTL on FRIDA ranges from 4.31\% to 13.79\% points. In terms of cumulative mAP, these networks improve by 4.97\% to 11.64\% points by training on FRIDA. Considering the large number of bounding boxes in FRIDA, these margins correspond to thousands of correct matches between identities. It is impressive that training on Market-1501 using 750 identities and 9,928 bounding boxes is outperformed by training on FRIDA with only 10 identities and less than 1,500 bounding boxes. This suggests that for an effective PRID on overhead fisheye images, having a higher variability of the viewpoint (including overhead) for each identity is more important than having more identities with less viewpoint variability. We note, however, that Pyramid is an exception to this observation. 
This seems to suggest that Pyramid is able to leverage a plurality of identities more effectively than viewpoint variability. 

%
%\mc{Here, I tried to address the Pyramid's performance drop, feel free to update as you wish}
%\mcrep{}{Unlike other networks, Pyramid performance worse when trained on FRIDA. This shows us that, for Pyramid, having more identities is more beneficial than having higher variability of appearance for each person.}

Table~\ref{tab:geo_segment_results_cross_validation} shows the performance of the geometry-based PRID algorithms for each segment of FRIDA. Clearly, all 4 algorithms did extremely well on segment 1 in which people are spread out fairly uniformly in the room and are never very close to each other. On the other hand, in segment 3 people stand very close to each other posing difficulties for location-based matching, resulting in the lowest performance among all segments. As expected, the algorithm based on the PPD distance metric (single query location mapping using an average person's height) achieves the lowest performance among the four algorithms. Algorithms using a range of person's height perform better with the CBD-based algorithm achieving the best performance in terms of the cumulative QMS (93.11\%) and cumulative mAP (96.97\%). This is consistent with observations in \cite{josh}.

For a fair comparison between appearance-based and geometry-based methods, we focus on the cumulative QMS and cumulative mAP values in Tables~\ref{tab:Market_1501_res_cross_validation}--\ref{tab:geo_segment_results_cross_validation}, computed over all FRIDA segments and accumulated over all camera pairs. 
It is clear that each geometry-based algorithm outperforms all appearance-based methods by a huge margin. However, it should be noted that in geometry-based algorithms a careful calibration must be performed once for each camera model (intrinsic parameters) prior to deployment. During installation cameras must be level-mounted and their extrinsic parameters (camera installation height, distance and rotation between cameras) must be measured or calibrated, a process known as system commissioning. Since this information may not be available in some scenarios, geometry-based algorithms would not be applicable in those cases. On the other hand, the appearance-based methods do not require any information about room setup and are easier to deploy in new scenarios.
% Even though the numbers for cam2+cam3 are low, don't forget the fact that we are working with ground truth bounding boxes right now. If we were to use an overhead fisheye people detection algorithm, it will miss the very tiny people who appear way further from the cameras, thus, they will not need to be re-identified from the perspective of double-counting the people.

%-------------------------------------------------------------------------------

\vspace{-1ex}
\section{Conclusions}

We introduced FRIDA, the first image dataset for person re-identification from overhead fisheye cameras. The dataset is unique not only for the camera type used but also for their overlapping fields of views that is often encountered when counting people in large spaces. This leads to a new type of PRID - matching of people ``seen'' by different cameras at the same time.

%In this paper, we introduced FRIDA, the first PRID dataset captured with overhead fisheye cameras. It is also unique in terms of its problem statement where people are re-identified between frames which are captured at the same instant with cameras which have overlapping FOV.

We evaluated the performance of 10 state-of-the-art PRID algorithms on FRIDA. Six of these algorithms were CNN appearance-based methods while four algorithms were based on geometry. The geometry-based algorithms performed significantly better than the appearance-based methods. The best-performing geometry-based method reaches almost 97\% in cumulative mAP, computed across all dataset segments and camera pairs. This is close to a perfect re-identification. Only in high-density scenarios (people close to each other causing severe occlusions), does its performance drop to about 95\%. However, geometry-based algorithms require calibration of each camera type used 
%(intrinsic parameters)
and additional measurements for each camera layout.

The appearance-based methods do not perform quite as well, even when trained on FRIDA, with the best one achieving below 83\% in cumulative mAP. This suggests there is much space for improvement in appearance-based methods. On the other hand, they require no calibration or measurements, and can be applied ``out of the box''.

We demonstrated that training CNN-based PRID methods on fisheye images improves performance when testing on fisheye images, which is not surprising. However, more research is needed to achieve {\it reliable} fisheye PRID in challenging scenarios (occlusions, high density of people). We hope FRIDA will inspire more research in this direction, and also serve as a benchmark for people detection, tracking and video-based PRID studies based on overhead fisheye cameras.

\vspace{-1ex}

%are also proven to be far from solving the overhead fisheye PRID with overlapping FOV. \mcrep{}{However, training the existing algorithms on FRIDA improved the results significantly and proven to be a step in the right direction to solve this problem.} Clearly, more research is needed in this field to transfer/adapt the traditional PRID methods to fisheye cameras. We are hoping FRIDA will inspire more research in this direction. Moreover, for future research, FRIDA can also serve as a benchmark dataset for people detection, people tracking and video-based PRID studies.

% In this paper, we formulated the problem as a image-based PRID not video-based PRID. So, we benchmarked the image-based algorithms.  \mcrep{}{It is important to acknowledge that FRIDA can be treated as image-based or video-based PRID dataset. In this paper, we treat it as image-based PRID and benchmark it accordingly.}

% %------------------------------------------------------------------------ 
% \section{Final copy}

% You must include your signed IEEE copyright release form when you submit 
% your finished paper. We MUST have this form before your paper can be 
% published in the proceedings.

{\small
\bibliographystyle{ieee}
\bibliography{egbib}
}

\end{document}